\documentclass[11pt]{article}

\usepackage[letterpaper,margin=1in]{geometry}

\usepackage[T1]{fontenc}
\usepackage{amsmath,amssymb,amsfonts}
\usepackage{amsthm}
\usepackage{graphicx}

\usepackage{algorithmic}
\usepackage{booktabs}
\usepackage{xcolor}
\usepackage{mathtools}
\usepackage{microtype}
\usepackage{url}
\usepackage[hidelinks]{hyperref}

\newcommand{\code}{\textbf}

\newcommand{\score}{\operatorname{score}}

\begin{document}

\title{Monitoring Neural Training with Topology: A Footprint-Predictable Collapse Index}
\author{Alexander Kalinowski\\SUNY Empire}
\date{}
\maketitle

\begin{abstract}
Representational collapse, where embeddings become anisotropic and lose multi-scale structure, can erode downstream performance long before performance metrics react.
We propose an online, topology-aware monitor for evolving neural representations that couples Modular Morse Homology Maintenance (MMHM) with a composite Collapse Index (CI).
Instead of rebuilding complexes each epoch, we apply sparse edits at a fixed scale and maintain a discrete Morse matching, yielding fast, incremental updates.
Across LLM fine-tuning and temporal KGE training, CI provides a low-latency early-warning signal suitable for in-training interventions.
Code and experimental scripts will be released publicly.
\end{abstract}


\section{Introduction}

Modern deep neural networks can drift into representational collapse: high-dimensional embeddings become anisotropic, lose multi-dimensional cluster structure, and crush topological cycles and holes, all features that support generalization.
One approach to detecting such a collapse is via measuring topological invariants of embedding layers and their evolution over model training epochs.
Topological invariants are well suited to expose this phenomenon because they summarize shape rather than individual coordinates.

In this paper, we develop an incremental topological monitor for evolving representation spaces.
Because recomputing topology per layer and epoch is costly, we use Modular Morse Homology Maintenance (MMHM) to maintain exact homology under sparse, fixed-scale edits.
We implement the MMHM-style local maintenance procedure independently from the public description in~\cite{ouyang2025morsebased}, and adapt it for training-time monitoring and an online collapse index (CI)  rather than for standalone topology maintenance.
The resulting CI can warn of structural degeneration in advance of task metrics and provides an avenue for actionable training decisions (early stop, LR schedule tweaks, data mix adjustments) informed by representational deformation in conjunction with task performance.
Our contributions are as follows:

\begin{enumerate}
    \item an application of an incremental MMHM update rule that preserves a valid Morse matching and critical cells to sparse neural embedding perturbations,
    \item a CI that combines computed Betti numbers, critical cell churn, and touched boundary ranks into a single metric designed to detect precursors to accuracy and calibration degradation, and
    \item empirical evidence on fine-tuning large language model (LLM) and training temporal knowledge graph (TKGE) embeddings, showing early-warning behavior ahead of accuracy drops and calibration drift.
\end{enumerate}

\section{Background}\label{background}

We introduce simplicial complexes and homology at a pragmatic level, then outline discrete Morse theory-- the engine behind our incremental updates, and close with a discussion on representational collapse.

Given an embedding layer $X^{(t)} \in \mathbb{R}^{N \times d}$ at epoch $t$, a \textit{simplicial complex} $K^{(t)}$ is a collection of vertices, edges, filled triangles, etc., that satisfies two properties: (1) $K^{(t)}$ is closed under taking faces, and (2) the non-empty intersection of any two simplices is a face of both simplices.
We construct $K^{(t)}$ from a 1-skeleton, or proximity graph based on $k$-NN neighborhoods, designed to capture adjacency structure in the embedding space, forgoing exact coordinate measures.
Given $K^{(t)}$, \textit{homology} captures the connected components, cycles, voids, and higher dimensional holes of the complex.
To calculate homology, we form a \emph{chain complex} on $K^{(t)}$ which consists of a sequence of vector spaces $C_0, C_1, \dots$ over a field $\mathbb{F}$ connected by linear boundary maps $\partial_n:C_n \to C_{n-1}$ such that $\text{im}(\partial_{n+1})\subset \text{ker}(\partial_{n})$.
In particular, the vector space $C_n$ has a basis consisting of $n$-simplices contained in $K^{(t)}$.
The $n^{th}$ homology group of $K^{(t)}$ is defined to be $H_n(K^{(t)}):=\text{ker}(\partial_{n})/\text{im}(\partial_{n+1})$.
The \emph{rank} of the $n^{th}$ homology group is the $n^{th}$ \emph{Betti number} $\beta_n$.
The zeroth, first, and second Betti numbers of $K^{(t)}$ provide the number of connected components, cycles, and voids respectively.
Homology can be computed using standard linear algebra techniques; we refer the reader to~\cite{edelsbrunner2010computational}.

\emph{Morse theory} is a way of studying the topology of a complex via real-valued functions on the complex.
\emph{Discrete Morse theory} equips a simplicial complex $K^{(t)}$ with a partial matching between incident cells (a $d$-simplex and one of its $(d-1)$-faces)~\cite{FORMAN199890}.
Cancellations of paired cells compress the boundary matrices without changing homotopy type~\cite{scoville2019discrete}.
In MMHM, this is exploited in two ways: a one-time discrete-Morse compression shrinks boundary matrices before any linear algebra, and after local edits, we repair the matching only inside affected local neighborhoods, re-establishing acyclicity and sparsity so that only touched columns $\mathcal{T}_d^{(t)}$ are re-reduced.
This locality preserves exact Betti numbers while keeping per-epoch work proportional to the touched-column footprint rather than to the size of the full complex, which is crucial under frequent, small embedding updates.

\subsection{Representational Collapse}

We introduce a complementary definition for representational collapse.
We say the representations \textit{completely collapse} at epoch $t$ if the variance along each coordinate falls below a small threshold, meaning the encoder outputs near constant vectors, as discussed in~\cite{complete_collapse}.
The representations \textit{dimensionally collapse} when most of the weight concentrates in a few principal directions, i.e., when a large fraction of the variance can be explained by a small $k \ll d$.
In our work, we employ the IsoScore measure of~\cite{rudman-etal-2022-isoscore}, further defined in Section~\ref{sec:hyp}.
Given a simplicial complex $K^{(t)}$ built on the embedding space, \textit{topological collapse} can be detected through shrinking Betti numbers and increasing critical cell churn in the associated Morse complex, indicating loss of alternative paths and voids in the geometry.
Our online monitoring CI aims to capture such topological collapse.


\section{Method: Incremental MMHM for Evolving Embedding Spaces}

Our objective is an online topological monitor with bounded, predictable latency per training epoch. 
We fix a single neighborhood scale and update one target complex across time, so small embedding drifts do not trigger wholesale recomputation. 
We build on MMHM to maintain homology under sparse edits, and introduce CI as a training-time collapse monitor.
We begin with a discrete Morse compression to a critical-cell complex, then, at each epoch, inject only local edits around points that moved most, repair the matching locally, and re-reduce the impacted boundary-matrix columns. 
This design keeps the number of column operations proportional to the touched footprint rather than the size of the entire complex, while preserving exact homology of the edited complex. 
We further expose internal maintenance signals, detailing how much and where linear algebraic work was done, which form inputs to a composite Collapse Index (CI) designed to trigger prior to conventional metrics deteriorate.
We target $\beta_0, \beta_1, \beta_2$ as these are informative about embedding space fragmentation ($\beta_0$) and void structure ($\beta_1, \beta_2$) and robust sample sizes required for embedding spaces.
The complex is a sparse clique-completion over a symmetric $k$-NN graph, with $k$ chosen to stabilize degrees (see Section~\ref{sec:hyp}). 
Each epoch, we refresh only simplices whose faces intersect the edited 1-skeleton (movers and their 1-hop neighbors).
A one-time greedy/coreduction pass produces an acyclic matching, where all subsequent steps operate on the compressed chain complex to minimize memory and column operations.

\subsection{Sparse and Incremental Complex Updates}

At epoch $t$ we identify a mover set $S(t)$ consisting of the top-$p\%$ points by displacement. 
The 1-skeleton is updated only inside the closed stars of $S(t)$: edges are added or removed under the fixed scale rule (mutual-kNN), and induced higher-dimensional simplices are refreshed near these edits.
The choice of $p$ controls sensitivity trade-offs: small $p$ yields low latency and detects large deformations, where a larger $p$ raises sensitivity to finer geometric changes at increased cost.
Following local skeleton edits, we repair the discrete Morse matching by re-pairing cells along short gradient paths confined to affected stars. 
We also track a per-dimension footprint $B^{(t)}_d$, the fraction of columns 
that were touched during repair, which both upper-bounds algebraic work and serves as a signal in our collapse index. 
The recompression rebuilds the matching on the current complex and resets locality constants.
Incremental reductions proceed only on columns flagged by the edit-and-repair phase. 
Since touched sets scale with $|S(t)|$ and local degree, the algebraic work is proportional to the footprint rather than the global complex size, yielding predictable per-epoch latency while preserving exact Betti numbers for the edited complex. 
Under contraction, the complex densifies: $\beta_0$ decreases while churn and footprint rise, with Betti shrinkage acting as a confirmatory signal.

\subsection{Collapse Signals}

To serve our goal of detecting early representation collapse during training, we seek signals that react quickly to local geometric changes and remain inexpensive under incremental updates.
These can be viewed as a weighted sum of topological reactivity against representation stability.
As explained above, our methodology measures $\beta_0, \beta_1, \beta_2$ through each evolution of the embedding space for evaluating topological changes.
Additionally, as the simplex underlying the embedding space changes according to our update policy, we can monitor changes to the discrete Morse matching. 
We first introduce the idea of critical cell churn $\chi$, intended to monitor representation instability.
Given that a discrete Morse matching encodes combinatorial cancellations, small, localized moves cause many re‑pairings and unpairings, making the complex structurally unstable.
Critical cell churn $\chi$ captures this directly: the more there is volatility in pairings and unpairings, the less stable the homological structure.

Even if churn is moderate, cycles can be brittle.
We introduce a secondary metric $R$, the cycle fragility, defined as the smallest local edit radius that cancels sampled persistent 1-cycles. 
For efficiency, we estimate this radius by the hop distance from each sampled cycle to the nearest touched edge, and report the median across samples.
We initialize a radius cap, based on the number of vertices and directed edges in the 1-skeleton as estimated when $t=0$.
The metric $R$ then represents the median hop-radius $(0, 1, 2, \ldots, r_{\text{cap}})$ from each sampled cycle to the nearest touched edge this epoch.
When $R = 0$, this is an indication that the homology is very fragile as a large proportion of the sampled cycles are incident to a touched edge, so small edits can lead to cancellations. 
As $R$ increases, cancellations are less likely, increasing stability of the homology.

Finally, we track the boundary footprint $B$, which measures how much of the boundary algebra was touched in order to keep the complex consistent after local edits. 
Intuitively, $B$ is a locality-aware ``work tracker”; when neighborhoods flip or matchings become unstable, more columns are touched and $B$ rises. 
Thus, $B$ serves a dual role-- a signal that lights up as topology becomes precarious, and a cost proxy as incremental run-time scales with the fraction of touched columns.
In regimes with small drifts, edits remain confined and $B$ stays low.
As collapse sets in, $B$ grows, indicating that the homological structure is becoming expensive to maintain.
%

The above metrics are complementary to traditional spectral cues. 
Spectral anisotropy metrics (e.g., IsoScore) can flag dimensional collapse, but they do not reveal which topological features are precarious.

We formalize these notions in the following definitions.
Let C(t) be the set of critical cells under matching $\mathcal{M}(t)$. 
Define $$\chi(t) = |C(t) \triangle C(t - 1)| / |C(t - 1)|$$ (symmetric‑difference rate) as the critical cell churn over each measurement epoch $t$.
For each persistent 1-cycle $\gamma$, represented by a minimal set of critical 1-cells, we can attempt cancellations within a small neighborhood of radius $r$.
Let $r^*(\gamma)$ be the smallest edit radius that kills $\gamma$.
We define 
$$R(t) = \operatorname{median}_\gamma r^*(\gamma)$$ 
as the median fragility radius over sampled cycles at epoch $t$.
For an approximation, one may bound $r$ by a small constant and report the fraction of $\gamma$ killed.

Let $K^{(t)}$ be the simplicial complex at epoch $t$. For $d \geq 1$, let 
    \begin{itemize}
        \item $\partial_d^{(t)} \in \mathbb{Z}_p^{n_{(d-1)^{(t)}} \times n_d^{(t)}}$ be the $d$-dimensional boundary matrix, 
        \item $S_d^{(t)}$ be the set of $d$-simplices with applied changes (added/removed/incident face changes) between $t-1$ and $t$, and
        \item $T_d^{(t)}$ be the set of $d$-columns affected by MMHM during the incremental update. 
    \end{itemize}
    We have $S_d^{(t)} \subseteq T_d^{(t)}$ as there can be extra columns perturbed during local eliminations or repairs.
    We then define the boundary footprint $B_d(t)$ as 
    $$B_d(t) = \frac{|T_d^{(t)}|}{n_d^{(t)}} \in [0, 1],$$
    representing the fraction of $d$-columns touched.

\subsection{Collapse Index}

Utilizing the features defined above, we aggregate to a single metric, the Collapse Index (CI) for tracking and predicting representation collapse.
The collapse index (CI) combines shape change ($\Delta \beta_k$), instability ($\chi$), fragility ($R$), and update footprint ($B$) to produce an early‑warning signal tied to homological structure.
Formally, we have
\begin{align}
    CI^{(t)} = 
w_0 \Delta \beta_0(t) +
w_1 \Delta \beta_1(t) &+
w_2 \Delta \beta_2(t) 
+ w_c \chi (t) \\
&+w_r \phi(R(t)) +
w_b B(t) \nonumber
\end{align}
where each score is z-scored per run and $\phi(R) = -R$.
Additionally, we compute $\Delta \beta_i(t) = \beta_i(t-1) - \beta_i(t)$, so $\Delta \beta_i(t)$ corresponds to shrinkage of $k$-dimensional features.
Optional smoothing using 
$$CI_\alpha (t) = \alpha CI_\alpha(t-1) + (1 - \alpha) CI(t)$$
is then applied for smoothing volatility over epoch updates.

In our work, we intuitively set the weights $w_i = 0.05$, $w_c=0.3$, $w_r=0.4$, and $w_b = 0.15$, with exponential smoothing $\alpha = 0.2$.
Fragility and churn carry the most weight; fragility is likely the more predictive feature, but is often harder to detect, only surfacing right before collapse actually happens.
The churn $\chi$ is assigned the next highest weight, and is independent from $B$, as it tracks pivot differences in the linear algebra, providing a complementary metric.
The footprint $B$ is a strong leading indicator of collapse due to the tracking of large, structural edits.
When the embedding space is on the verge of collapse, the localized algebra in MMHM touches more critical columns.
The weights for deltas in Betti numbers are smallest, as they serve as a confirmatory variable of collapse. 
Small jitters in the points moved between epoch can surface changes in the Betti numbers that are less robust to detecting actual collapse, thus we weight them the smallest.

\section{Experimental Design}

We focus on the fine-tuning of LLMs as well as the full training cycle of temporal knowledge graphs across multiple tasks and datasets as a testbed for the incremental MMHM methodology and newly introduced collapse index metric.
All topology quantities are maintained incrementally across epochs; only the local neighborhood induced by the mover set is refreshed, while the remainder of the complex is unchanged.

\subsection{LLMs Experiments}~\label{sec:llm-design}

The STS-B dataset evaluates sentence–sentence semantic similarity as a regression task. 
Each example consists of two sentences drawn from multiple sources (news, forums) and a human annotated similarity score in $[0,5]$ indicating how semantically equivalent the pair is. 
We fine-tune an encoder with a single linear head to predict the continuous score, minimizing the mean squared error against the human annotated scores.
We selected three common pre-trained language models for our fine tuning experiments.

\begin{itemize}
    \item \code{bert-base}: a baseline model with model weights retrieved from Hugging Face (dimension $d=768$)
    \item \code{sbert-base}: a model fine-tuned to relevance of the task (dimension $d=768$)
    \item \code{allMini-base}: a smaller model for comparison of layer dimensionalities (dimension $d=384$)
\end{itemize}

For each model, we construct per-epoch embedding snapshots and a fixed-scale simplicial complex that is subsequently updated by MMHM.
For the selected LLMs, inputs are tokenized using the model's native tokenizer.
Representations are taken either as $[CLS]$ vectors or mean-pooled embeddings (as specified in each experiment) with L2 normalization.
For our selected LLMs, we cache layers $[0,3,6,9]$ to capture an array of depths across model training (all have 12 layers), as motivated and informed by~\cite{bert-classical-pipeline, bert-structure, bertology} to access various structural aspects of language representations.

\subsection{TKGE Experiments}\label{sec:tkge-design}

A temporal knowledge graph consists of quadruples $\langle s,p,o,t\rangle$ stating that relation $p$ holds from subject $s$ to object $o$ at time $t$. 
Temporal KGE models embed entities, relations, and time, and define a time-conditioned $\score(s,p,o,t)$ used to rank candidate links. 
We evaluate on ICEWS14, ICEWS05-15, Wikidata-12k, and YAGO. 

We study three families that cover complementary ways of injecting time: additive translation, phase rotation, and multiplicative modulation (\code{TransE-TE}, \code{RotatE-TE}, and \code{ComplEx-TE}, respectively)~\cite{transe-te, lacroix2018canonicaltensordecompositionknowledge, sun2019rotateknowledgegraphembedding}.
For models with complex-valued or probabilistic parameters, we build a Euclidean representation by taking the real concatenation in the complex case or the mean vector in the probabilistic case. 
This allows a uniform fixed-scale graph construction for our topology pipeline without adding extra model-specific adapters. 
For all models we sweep embedding dimension $d\in\{50,100,200\}$; lower $d$ emphasizes efficiency and tends to constrain expressivity, while higher $d$ increases expressibility for modeling asymmetric patterns and temporal interactions, but it also raises the risk of overfitting and increases compute. 
Reporting across these dimensions lets us separate architectural effects from capacity effects in both link prediction and representational collapse.
As the tasks are independent of the architectures, this provides a contrast to the fine-tuning experiments of Section~\ref{sec:llm-design}.

\subsection{Hyperparameters and Metrics}\label{sec:hyp}

To keep tracking of CI consistent across models, we also keep our hyperparameters consistent across training runs.
For all classes of models, we fine-tune over 30 epochs, performing runs with three random seeds.
We select a constant learning rate of $1e^{-3}$ without the use of a scheduler; this controls the embedding space evolution, whereas schedulers may dilate the space as they hone in on and exploit particular features.
For LLMs, we keep the embedding model for each layer equal to those used in the model pre-training (as detailed in Section~\ref{sec:llm-design}).
Inputs are tokenized with the model’s native tokenizer; later representations are taken as either CLS vectors or mean-pooled token embeddings (as specified in each experiment) and L2-normalized. 
For TKGE models, we sweep over embedding dimensions $d \in \{50, 100, 200\}$; while these dimensions are significantly smaller than those used in the associated papers, they keep our compute footprint low and do not require additional hardware.
The number of negative samples $\eta$ is set to 50 for each model run.
While these selections do not reach the SOTA metrics reported in the papers, they keep runtime and memory low while generating the necessary embedding spaces for measurement.
We sweep over nearest neighbors in the range $k \in \{2, 4, 8, 16, 32\}$ and  $p \in \{0.05, 0.1, 0.2, 0.3, 0.4, 0.5\}$ for LLM models and $p \in \{0.05, 0.1, 0.2, 0.3\}$ for TKGE models due to density, as later explored.
For TKGE models, we measure filtered MRR/Hits@K ($K=1,5,10$); for LLM models we measure Pearson and Spearman correlations with the ground truth labels, as well as MSE.

We additionally test whether $CI^{(t)}$ derives its predictive power from the careful initial weighting or from the alignment of its features with collapse dynamics.
Concretely, we ablate the individual components included in the CI metric by recomputing with each individual weight set to zero.
In each case, the weights are redistributed (to sum to 1) when calculating the with/without lagged correlations of the target metric and the CI metrics.
This ablation provides evidence to support the motivation for our initial weights as well as insights into which CI pieces are best capturing the evolution of the underlying topology.

We finally compare our CI diagnostics to measures of spectral isotropy/anisotropy for comparison with topological reactivity. 
IsoScore is a scalar isotropy measure defined on a set of embeddings
\(x_1, \dots, x_n \in \mathbb{R}^d\) via their empirical covariance
matrix \(\Sigma\). Let \(\lambda_1, \dots, \lambda_d\) denote the
eigenvalues of \(\Sigma\), and define the normalized spectrum
\[
p_i \coloneqq \frac{\lambda_i}{\sum_{j=1}^d \lambda_j}, \qquad i = 1,\dots,d,
\]
and the uniform distribution on coordinates
$u_i = \frac{1}{d}, i = 1,\dots,d.$
IsoScore is then defined as a normalized distance between
\(p = (p_1,\dots,p_d)\) and \(u = (u_1,\dots,u_d)\), rescaled to lie in
\([0,1]\). 
In particular, \(\text{IsoScore} = 1\) if and only if all
\(\lambda_i\) are equal (maximally isotropic use of \(\mathbb{R}^d\)),
while \(\text{IsoScore} \to 0\) as the variance concentrates in a small number of directions (maximally anisotropic).

\subsection{Hypotheses}

We introduce the three main hypotheses we intend to explore experimentally: \textbf{H1} (early-warning) CI increases ahead of task degradation, evidenced by magnitude correlations at negative lags between CI and downstream metrics,
\textbf{H2} (sensitivity vs. budget) increasing the mover fraction $p$ increases sensitivity (larger $\chi$, lower $R$) but raises $B(t)$ and latency in a predictable manner, and \textbf{H3} (CI vs. isotropy) CI  will capture model degradation prior to IsoScore.

\section{Results and Analysis}\label{results}

We present the results of our experiments on LLMs and TKGEs in the following two sections, including ablation studies on the components of CI and results with respect to scaling of $k$ and $p$.
Unless otherwise noted, we report the mean metrics over three random seeds.

\subsection{LLM Results}\label{sec:llm-results}

\begin{center}
\begin{figure}[hbt!]
  \includegraphics[width=\linewidth]{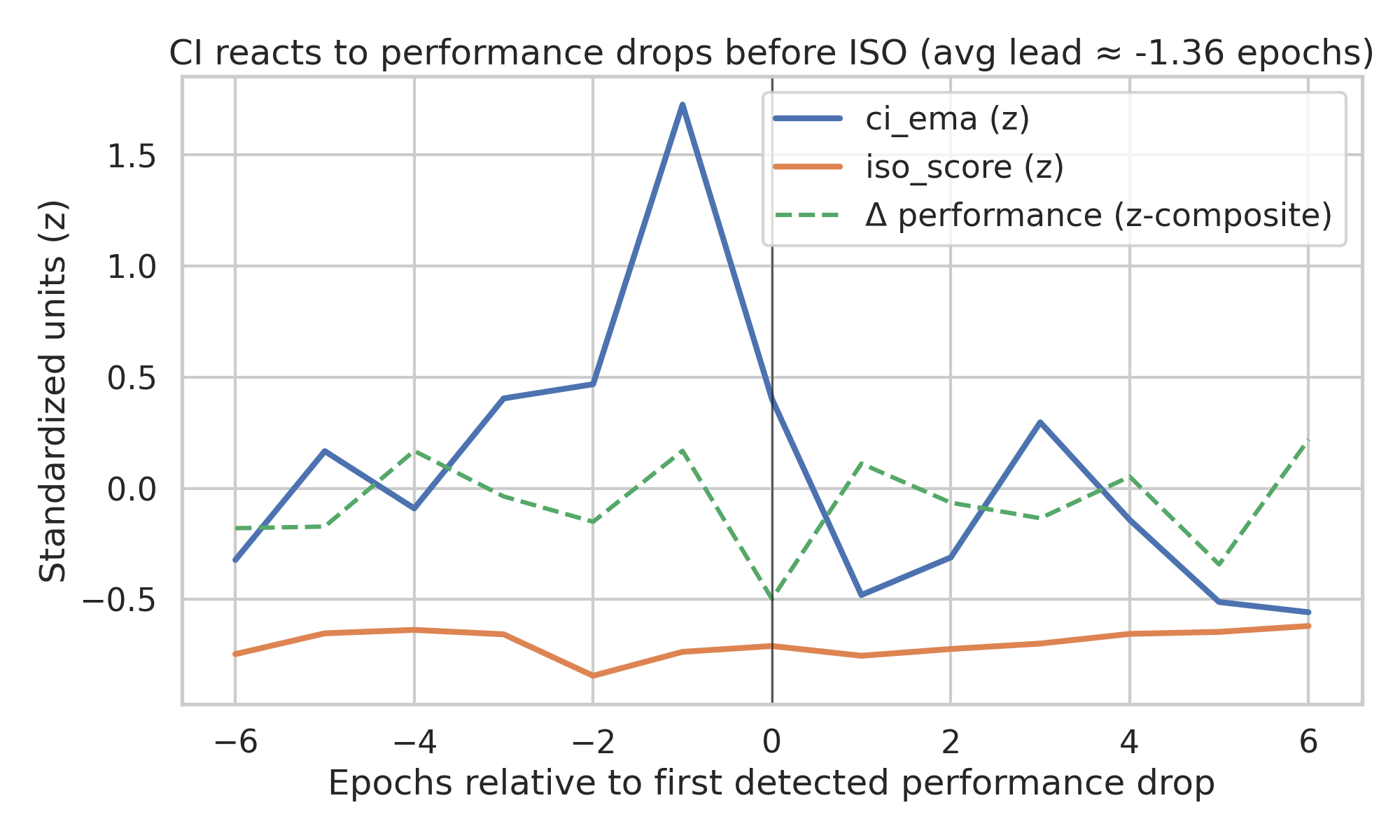}
\caption{A comparison of CI and IsoScore for leading detection of LLM model performance drops. Not only does CI give a greater magnitude (and more easily detectable) signal, but this signal occurs on average 1.36 epochs before the corresponding, weaker signal from IsoScore. Performance metrics reported here are a z-scored average of Pearson, Spearman, and MSE (z-composite).}
\label{fig:llm_ci_iso}
\end{figure}
\end{center}

We first address \textbf{H1}, as exhibited in Figure~\ref{fig:llm_ci_iso}.
Across all of our experiments, independent of the selection of hyperparameters $p$ and $k$, our CI metric, on average, reacts 1.4 epochs prior to our compared isotropy metric (IsoScore).
Even across diverse architectures and embedding dimensionality, the signal from CI, but tracking the underlying topology of the space and the evolution of changes over epochs, can provide an early warning sign of representational collapse and trigger interventions (LR changes, early stopping of parameter sweeps).

For \code{sbert-base} and \code{allMini-base}, CI reliably increases and leads to performance drops by approximately 4–5 epochs (Table~\ref{tab:llm-ci-summary}), with \code{sbert-base} showing the strongest effect leading performance.
For \code{bert-base}, CI is less consistent across $(p, k)$ settings, suggesting greater sensitivity to representation choice and hyperparameters in this model.
While average lead time is positive by approximately 3 epochs, this is influenced by a few outlier models for choices of $p$ and $k$, and thus the negative lag metrics actually exhibit positive correlations.
As we dissect the remaining results, we show that the best performing range for hyperparameters $p$ and $k$ are dependent on embedding dimensionality and can reverse this correlation.

We focus first on \code{sbert-base}, where the results are strongest (indicated by bold values in Table~\ref{tab:llm-ci-summary}).
At layer 6 (the closest layer to sentence representations), the maximum average lead time per experiment is 5.6 epochs, far in advance of model performance degradation and representational collapse.
This similarly holds for all models-- as layer 6 is arguably the critical layer to the task (sentence similarity scoring), we see CI react strongly at that point.
For \code{sbert-base}, the average negative lag correlation correspondingly aligns.
Overall, the top performing models (detailed in Table~\ref{tab:llm-best}) with respect to correlation between performance metrics and CI are all attributed to the architecture of \code{sbert-base}.
While we also note the top performance of alternate models, CI shows the strength of topological monitoring for task-specific models: it can `weed out' architectures prior to spending compute resources on fine-tuning models which already exhibit collapse, whether it be dimensional collapse or topological collapse.

\begin{table}[t]
\centering
\small
\caption{The top 5 best performing models for \code{sbert-base}, our leading performer with respect to CI, as well as the top two best performing models for \code{allMini-base} and \code{bert-base}, for discussion of selection of $p$ and $k$.}
\label{tab:llm-best}
\begin{tabular}{lccccc}
\toprule
Model & Layer & $p$ & $k$ & $\rho_{neg}$ & lead \\
\midrule
\code{sbert-base} & 6 & 0.20 & 32 & -0.91 & 6.00 \\
\code{sbert-base} & 3 & 0.40 & 32 & -0.90 & 6.00 \\
\code{sbert-base} & 6 & 0.05 & 32 & -0.90 & 6.00 \\
\code{sbert-base} & 9 & 0.30 & 32 & -0.89 & 6.00 \\
\code{sbert-base} & 3 & 0.30 & 32 & -0.89 & 1.00 \\
\midrule
\code{allmini-base} & 3 & 0.50 & 16 & -0.86 & 6.00 \\
\code{allmini-base} & 3 & 0.40 & 16 & -0.86 & 6.00 \\
\midrule
\code{bert-base} & 0 & 0.40 & 32 & -0.31 & 1.00 \\
\code{bert-base} & 0 & 0.30 & 32 & -0.28 & 1.00 \\
\bottomrule
\end{tabular}
\end{table}

We found that configuring $p$, while inherently linked to the boundary footprint $B$, can be limited to a small range, the selection of $k$ was dependent on the underlying dimensionality.
For models with greater dimensionality (\code{sbert-base, bert-base} where $ d=768$), there was a greater need to seed the initial 1-skeleton with a larger selection of $k$, typically 32 nearest neighbors.
Due to this increasing dimensionality, the smaller amount of principal components, and the quickly collapsing IsoScore, the initial snapshot of the embedding space requires additional simplices to be added to the MMHM machinery for more fine-grained tracking.
In smaller models, such as \code{allMini-base}, we are able to circumvent these issues due to the smaller initial dimension size ($d = 384$), decreasing initial sparsity in the 1-skeleton and allow for earlier signals, both in terms of per-layer collapse and lag-correlation.
In conjunction with this finding, \code{allMini-base} required more sampled points of the space $p$ in order to trigger these early warning signals.
We further explore the relationship between dimensionality $d$, nearest neighbors $k$, and points moved $p$ when assessing TKGE models, where full training regimes are run and dimensionality is controlled for, in Section~\ref{sec:tkge-results}.

\begin{table}[t]
\centering
\small
\caption{STS-B (LLM) early--warning summary by model and layer. More negative $\overline{\rho_{\text{neg}}}$ indicates stronger early warning (for higher-is-better STS-B metrics). $\overline{\text{lead}}$ is the mean best negative lag (epochs). Rightmost column: fraction of configurations where $|\rho_{\text{neg}}|>|\rho_{0}|$.}
\label{tab:llm-ci-summary}
\begin{tabular}{lcccc}
\toprule
Model & Layer & $\overline{\rho_{neg}}$ & $max\_lead$ & \%\,neg $>$ zero \\
\midrule
\code{allmini-base} & 0 & -0.15 & 4.03 & 66\% \\
\code{allmini-base} & 3 & \textbf{-0.43} & 4.63 & 66\% \\
\code{allmini-base} & 6 & -0.26 & \textbf{5.50} & 66\% \\
\code{allmini-base} & 9 & \textbf{-0.43} & 5.17 & 66\% \\
\addlinespace
\code{bert-base} & 0 & \textbf{0.31} & 2.83 & 52\% \\
\code{bert-base} & 3 & 0.47 & 2.70 & 52\% \\
\code{bert-base} & 6 & 0.59 & \textbf{3.20} & 52\% \\
\code{bert-base} & 9 & 0.58 & 3.13 & 52\% \\
\addlinespace
\code{sbert-base} & 0 & -0.06 & 5.37 & 63\% \\
\code{sbert-base} & 3 & -0.47 & 4.43 & 63\% \\
\code{sbert-base} & 6 & \textbf{-0.58} & \textbf{5.60} & 63\% \\
\code{sbert-base} & 9 & -0.45 & 4.83 & 63\% \\
\bottomrule
\end{tabular}
\end{table}

With regards to the information captured in the IsoScore, the metric tends to increase with CI and anti-correlate with task performance, aligning with the hypothesis that anisotropy/dimensional collapse accompanies topological fragility.
CI often moves first, then isotropy, then the task metric; pairing both together is useful for early stopping or LR scheduling.
This is depicted in Figure~\ref{fig:llm_ci_iso}, showing the lead time of CI to IsoScore for a composite performance metric (z-score normalization of Spearman, Pearson, and MSE).
As an average across all experiments, CI reacts approximately 1.4 epochs prior to IsoScore, validating the hypothesis of \textbf{H3}.

Looking at performance by each individual model, \code{sbert-base} has the strongest and most stable early-warning overall.
At layers 3 and 6 with top-p between 10–20\%, the best negative-lag Pearson/Spearman are the largest in magnitude with clean lead (1-3 epochs), and lagged-MSE improves over zero-lag.
The EMA further stabilizes lead detection and should be used as the primary signal.
The better performance of the CI signal when fine-tuning \code{sbert-base} is unsurprising as the model is explicitly designed for the STS-B task.
For the \code{bert-base} model, CI still leads metrics, but effect sizes are smaller and more variable across top-p.
The best signals are again in layers 3 to 6 and top-p 10–20\%, but there are more experimental results where the best negative-lag correlation is modest.
In \code{bert-base}, the geometry is less readable for sentence level performance and CI needs a bit more top-p budget to signal reliably.
For \code{allMini-base}, signals are present but attenuated. 
The best negative-lag correlations appear at layer 6 with top-p between 10\% and 20\%, but with shorter leads and smaller magnitudes than \code{sbert-base}. 
At 20\% top-p, the CI stays active but becomes less predictive due to copious editing throughout the model.
This aligns with the smaller dimensionality of this more compressed model.

\subsubsection{CI Component Ablation} Across all three LLMs and layers, removing the fragility term produces the largest drop in best negative-lag correlations ($\Delta \approx (-0.02, -0.04)$), followed by removal of churn and footprint. 
In contrast, removing any individual Betti delta has negligible or slightly positive effect ($\Delta \approx 0$, thus not reported), confirming that fragility and churn are the primary drivers of CI’s predictive power, with $B$ as a useful but secondary signal and $\Delta \beta$s acting mainly as a confirmatory term.
Table~\ref{tab:ci-ablate-ema-full} confirms that removing $R$ (no $R$) consistently harms CI the most, supporting our claim that fragility drives specificity.
Removing $\chi$ (no\_churn) is the second-most damaging ablation, consistent with its role as an early instability detector.

\begin{table}[t]
\centering
\small
\caption{Ablation $\Delta$ (mean best negative-lag) for \textbf{Spearman} using EMA CI.}
\label{tab:ci-ablate-ema-full}
\begin{tabular}{l r r r r r r r}
\toprule
Model & Layer & w/o $\chi$ & w/o $R$ & w/o $B$ \\
\midrule
\code{sbert-base}     & 0 & -0.011 & \textbf{-0.026} & -0.007 \\
\code{sbert-base}     & 3 & -0.015 & \textbf{-0.035} & -0.010 \\
\code{sbert-base}     & 6 & -0.017 & \textbf{-0.040} & -0.012 \\
\code{sbert-base}     & 9 & -0.013 & \textbf{-0.031} & -0.009 \\
\addlinespace
\code{bert-base}      & 0 & -0.008 & \textbf{-0.020} & -0.006 \\
\code{bert-base}      & 3 & -0.011 & \textbf{-0.028} & -0.008 \\
\code{bert-base}      & 6 & -0.013 & \textbf{-0.032} & -0.009 \\
\code{bert-base}      & 9 & -0.010 & \textbf{-0.025} & -0.007 \\
\addlinespace
\code{allMini-base }& 0 & -0.007 & \textbf{-0.016} & -0.004 \\
\code{allMini-base }& 3 & -0.009 & \textbf{-0.022} & -0.005 \\
\code{allMini-base }& 6 & -0.010 & \textbf{-0.025} & -0.006 \\
\code{allMini-base }& 9 & -0.008 & \textbf{-0.020} & -0.005 \\
\bottomrule
\end{tabular}
\end{table}

\subsubsection{Scalability}~\label{sec:llm-scale} We expect the boundary footprint to scale approximately linearly with the top-p\% moved points.
We verify this relationship holds by fitting a linear model between $B(t)$ (the aggregate footprint per epoch) and the size of $\mathcal{S}^{(t)}$ (the set of moved points).
Our findings show that such a linear model is well fit, with an adjusted $R^2 = 0.795$ across all LLM models, with similar results when the regression is performed on a per-model basis.
The model additionally yields a fit coefficient for top-p\% of 1.931.

\begin{center}
\begin{figure*}
\begin{tabular}{cc}
  \includegraphics[width=0.48\linewidth]{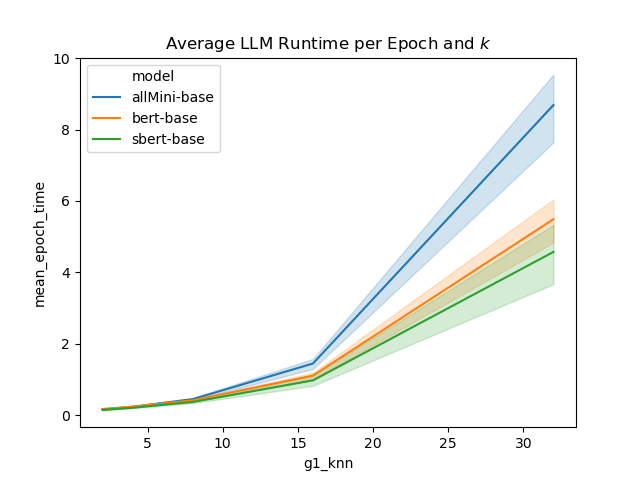} &
  \includegraphics[width=0.48\linewidth]{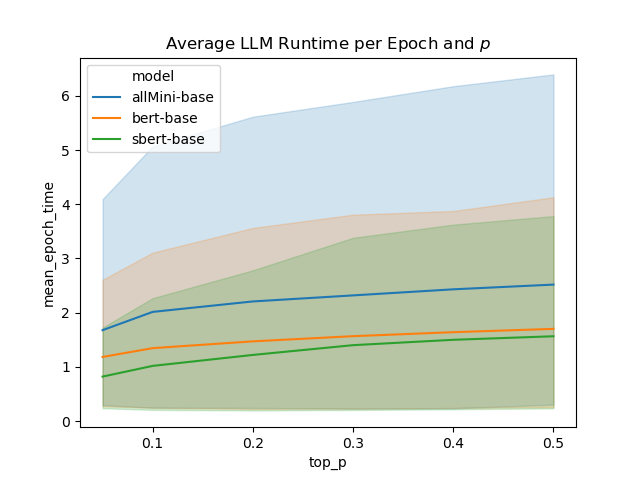} \\
  (a) Mean Runtime vs. kNN &
  (b) Mean Runtime vs. Top-p\% \\
\end{tabular}
\caption{A comparison of CI compute times (under the MMHM engine) for $p$ and $k$ across all epochs and layers for LLM experiments. }
\label{fig:llm_latency}
\end{figure*}
\end{center}

\subsection{TKGE Results}~\label{sec:tkge-results}

Overall, we find the results of full training regimes over TKGE models to be broadly consistent with the fine-tuning results of LLM fine-tuning, although with some caveats per each hypothesis.
Our hypothesis on scaling (\textbf{H2}) holds even stronger than the LLM results, with a linear relationship with $R^2 = 0.98$.
The majority of fits on hyperparameters $p$ and $k$ have $R^2 \geq 0.98$ and positive slopes, with only a small minority (mostly at $k=2$ on the smallest $d$) do not fit the same pattern.
This agrees with our assessment, not only in the dimensionality size mentioned in Section~\ref{sec:llm-results}, but subsequently with the relationship with the overall density of the knowledge graph, the average degree of the initial 1-skeleton, and the selection of $k$.

We additionally find that architecture is a driving factor in detecting collapse.
Leading with \code{Rotate-TE}, the model most expressive of potential relational patterns in a knowledge graph, CI is able to detect earlier collapse, on average approximately 3.6 epochs earlier.
As in the case of \code{sbert-base}, the more attuned the architecture to the task, the closer CI generalizes to detecting earlier warning signs; here, even across diverse datasets.
The next best is \code{ComplEx-TE}, with an average of lagged performance degradation at 2.8 epochs.

When comparing to IsoScore, the results are less dramatic than in the case of LLMs, due to the nature of full training versus fine-tuning, and the fact that the majority of the models trained do not collapse as severely as in the LLM experiments.
Overall, CI leads IsoScore in 30\% of our experiments, and is closely-tied with IsoScore in 40\% of experiments.
On a per-model basis, CI leads 3.1 epochs prior to performance degradation in \code{ComplEx-TE}, followed by \code{TransE-TE} with lead of 2.87 epochs, and finally \code{RotatE-TE} with a lead of 2.74 epochs. 
While these leads are shorter than those in our LLM experiments, model collapse is less prevalent.
When compared to IsoScore, CI is the early leading indicator when using \code{ComplEx-TE}, while showing only marginal leads (mean of 0.9 epochs) for the other models.
While there is slightly weaker evidence to support \textbf{H3} in our TKGE experiments, CI provides an alternate, yet complementary metric to traditional anisotropy scores.

\subsubsection{CI Component Ablation} 

In ablating each component of CI for the TKGE tasks, we again find that fragility plays the largest role, as demonstrated in Table~\ref{tab:tkge-ci-ablation-tkge-only}.
Removing fragility, on average, decreases the EMA CI measure by 0.42, meaning that nearly half of the component is based on this critical metric.
We also find that in most instances, fragility plays an increasingly important role as the dimension size grows.
The second next important component is churn, on average decreasing EMA CI by 0.26.
Again, in most instances, increasing dimensionality of the embedding layer increases the prevalence of critical cell churn.
Finally, removal of the footprint, on average, decreases the EMA CI signal by 0.11.
In this ablation, we see that the components defined in our CI play a more critical role than in LLM training.
We attribute this to full model training in the TKGE setting, rather than fine-tuning in the corresponding LLM experiments.
As the models start with a blind set of embedding layers, per-epoch changes are more dramatic, thus influencing churn, fragility, and boundary column edits.
This further highlights the utility of our CI metric as an early-warning indicator that can be computed during model training to prevent experiments heading toward degenerate models.

\begin{table}
\centering
\small
\caption{TKGE-only CI Ablation: Mean EMA CI (aggregated across datasets) for each model and dimension.}
\label{tab:tkge-ci-ablation-tkge-only}
\begin{tabular}{lcrrr}
\toprule
Model & Dimension & w/o $\chi$ & w/o $R$ & w/o $B$ \\
\midrule
\code{complex\_te} & 50 & -0.2235 & \textbf{-0.3500} & -0.0916 \\
\code{complex\_te} & 100 & -0.3055 & \textbf{-0.4807} & -0.1246 \\
\code{complex\_te} & 200 & -0.2622 & \textbf{-0.4123} & -0.1072 \\
\code{rotate\_te} & 50 & -0.2374 & \textbf{-0.3751} & -0.0968 \\
\code{rotate\_te} & 100 & -0.2535 & \textbf{-0.3969} & -0.1039 \\
\code{rotate\_te} & 200 & -0.4134 & \textbf{-0.6474} & -0.1695 \\
\code{transe\_te} & 50 & -0.2054 & \textbf{-0.3192} & -0.0846 \\
\code{transe\_te} & 100 & -0.2461 & \textbf{-0.3895} & -0.1007 \\
\code{transe\_te} & 200 & -0.2682 & \textbf{-0.4303} & -0.1099 \\
\bottomrule
\end{tabular}
\end{table}

\subsubsection{Scalability}~\label{sec:tkge-scale} The same scaling laws as in LLM models hold for TKGE models, with a caveat with respect to the underlying knowledge graph density (See Figure~\ref{fig:tkge_runtime} in Appendix).
As density of the graph increases, as does the density of the embedding space for both entities and predicates.
This drives an increase in average degree of the 1-skeleton as compared to LLM models.

\section{Related Work}

Related work on vineyards/zigzag/streaming PH focuses on updating persistence under global filtrations, often revisiting many columns per update. In contrast, we maintain a fixed-scale complex and restrict updates to local stars of top-p\% movers, enabling footprint-bounded incremental reductions suited to training-time monitoring.

\section{Conclusion}

In this work we introduce a topology-aware monitoring pipeline and metric, the collapse index (CI).
We couple Modular Morse Homology Maintenance (MMHM) with composite measures of Betti numbers, critical cell churn, fragility of 1-cycles, and boundary matrix operations to efficiently compute collapse signals using a small percentage of point updates during neural training.
This efficiency allows for CI to be computed during model training, and is a leading indicator of model collapse when compared to performance metrics.
While this lead is strong for task-aware LLM fine-tuning, it serves as a complementary metric to anisotropy metrics (IsoScore) when applied to TKGE ground-up training.
A minor limitation of CI is sensitivity to scale ($k$) and the choice of representation layer.

\clearpage
\bibliographystyle{plain}
\bibliography{mmhm}

\appendix
\section{Appendix}

\begin{center}
\begin{figure*}
\begin{tabular}{cc}
  \includegraphics[width=0.48\linewidth]{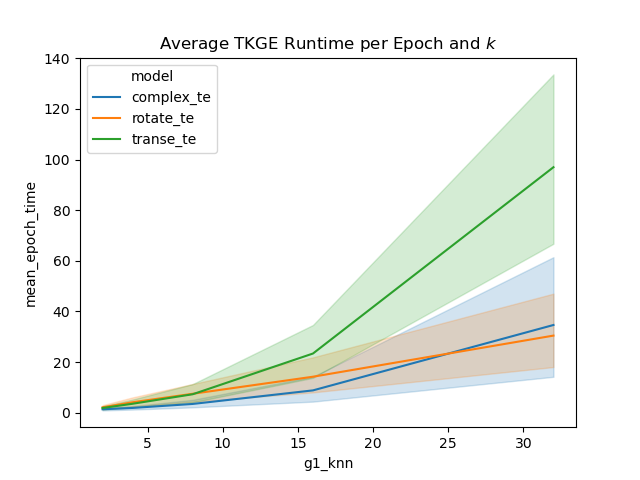} &
  \includegraphics[width=0.48\linewidth]{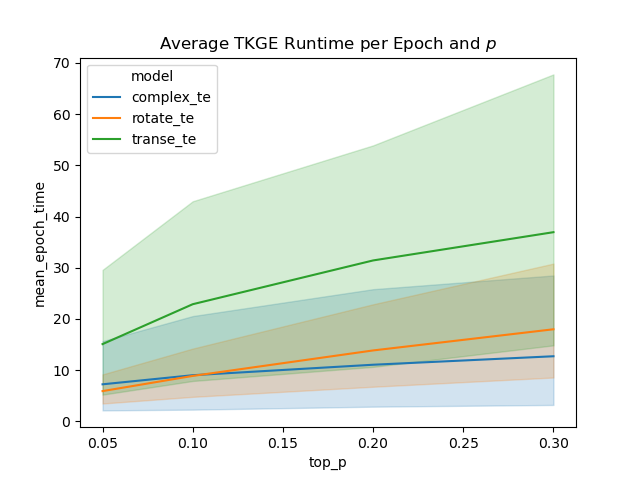} \\
  (a) Mean Runtime vs. kNN &
  (b) Mean Runtime vs. Top-p\% \\
\end{tabular}
\caption{A comparison of CI compute times (under the MMHM engine) for $p$ and $k$ across all epochs, datasets, and dimensions for TKGE experiments. }
\label{fig:tkge_runtime}
\end{figure*}
\end{center}

\end{document}